\documentclass[letterpaper, 10 pt, conference]{ieeeconf}  % Comment this line out if you need a4paper

\IEEEoverridecommandlockouts                              % This command is only needed if 
\usepackage{epsfig} % for postscript graphics files
\usepackage{times} % assumes new font selection scheme installed
\usepackage{amsmath} % assumes amsmath package installed
\usepackage{amssymb}  % assumes amsmath package installed
\usepackage{color}
\usepackage{xcolor}
\usepackage{graphicx}
\usepackage{url} 
\usepackage[format=plain,labelsep=period]{caption}
\usepackage{wrapfig}
\usepackage[list=on,listformat=simple]{subcaption}
\usepackage{makecell}
\usepackage{float}
\usepackage{booktabs}
\floatstyle{plaintop}
\restylefloat{table}
\usepackage{mathtools}
\usepackage{hyperref}
\usepackage{cite}
\usepackage{graphicx}
\usepackage{multirow}
\usepackage{colortbl, booktabs}

\usepackage{preambles}

\usepackage[capitalise]{cleveref}

\title{\LARGE \bf
Learning Navigation from Multimodal Demonstrations with \protect\\ Geometric Candidate Modes}
\author{Chengyang Peng$^{1}$, Zhihao Zhang$^{2}$, Upama Pant$^{1}$, Yiming Wang$^{1}$, Keith A. Redmill$^{2}$, and Ayonga Hereid$^{1}$% <-this % stops a space
\thanks{$^{1}$Mechanical and Aerospace Engineering, The Ohio State University, Columbus, OH, USA. {\tt\footnotesize (peng.947, pant.46, wang.16925, hereid.1)@osu.edu.}}%
\thanks{$^{2}$Electrical and Computer Engineering, The Ohio State University, Columbus, OH, USA. {\tt\footnotesize (zhang.11606, redmill.1)@osu.edu.}}%
}
\begin{document}

\maketitle
\thispagestyle{empty}
\pagestyle{empty}

\begin{abstract}
Offline navigation demonstrations are multimodal, because one observation often allows several valid actions, such as passing an obstacle on either side. Behavior cloning averages these actions, potentially leading to unsafe routes. Multimodal policies must discover the alternatives from the data, but similar states can lead to different routes whose current actions overlap. In navigation, the alternatives have a geometric cause and can be computed from one observation. We propose a hierarchical neural-network policy that computes candidate modes from local geometry and learns from demonstrations which mode to follow and how. A high-level policy selects a mode using the goal and its previous choice among a set of candidate modes, and a low-level policy turns it into an action.
In simulation, the policy succeeds in over 95\% of seen and unseen test cases, compared with at most 43\% and 20\% for four offline learning baselines and 79\% and 67\% for a heuristic that uses the same modes without learning. The simulation-trained policy also transfers to a LimX Tron 1 bipedal robot.
\end{abstract}

\section{Introduction}
\label{sec:intro}

\begin{figure}
\vspace{2mm}
\centering
    \begin{subfigure}[b]{0.49\columnwidth}
         \centering
         \includegraphics[width=\textwidth]{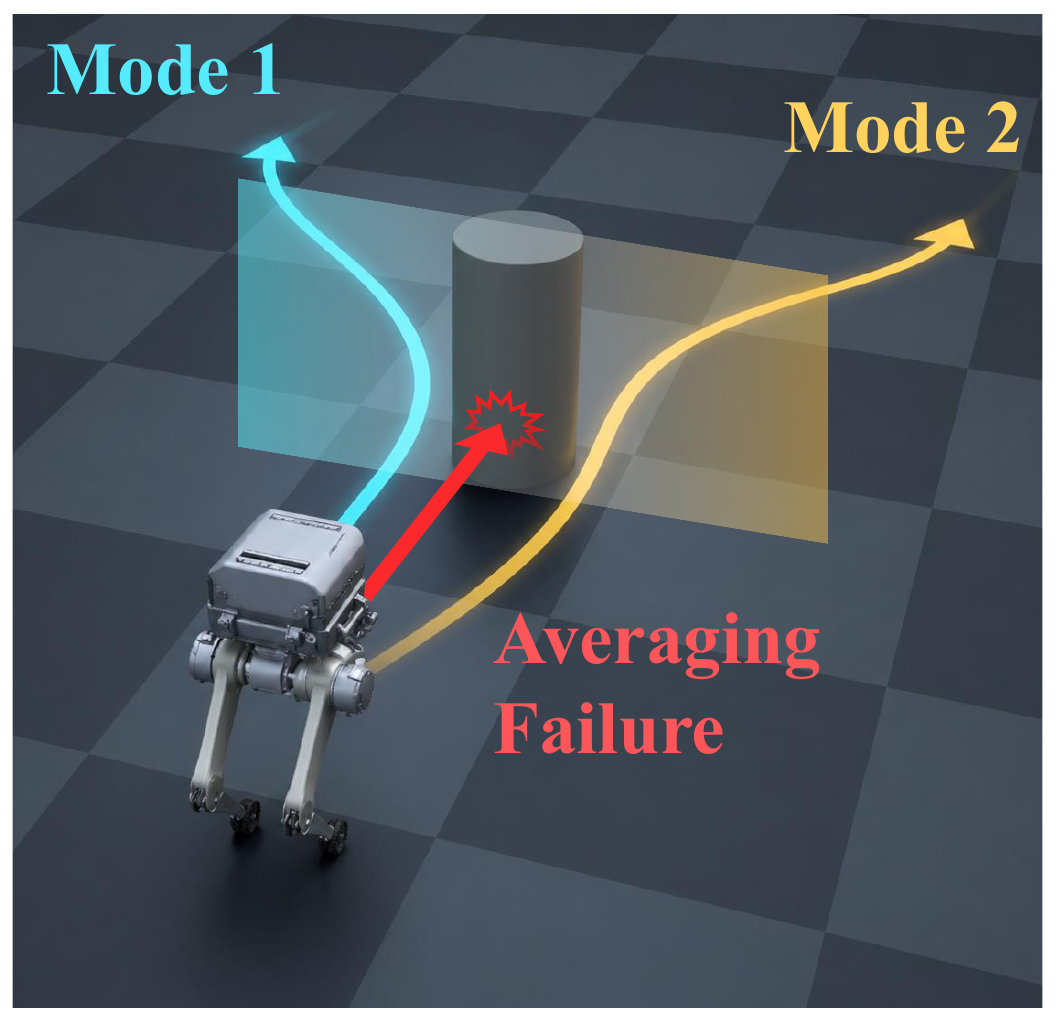}
         \caption{}
         \label{fig:scenarios}
     \end{subfigure}
     \hfill
     \begin{subfigure}[b]{0.49\columnwidth}
         \centering
         \includegraphics[width=\textwidth]{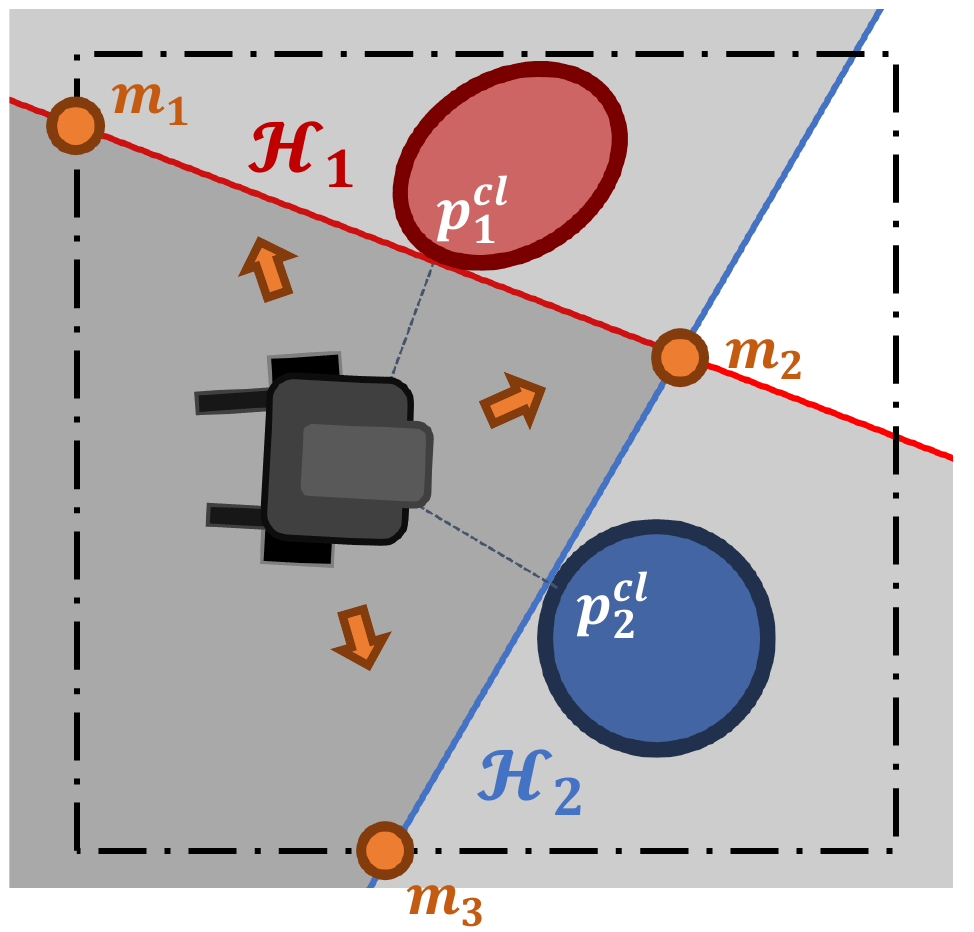}
         \caption{}
         \label{fig:modeling}
     \end{subfigure}
\caption{\small{(a). Averaging two feasible avoidance trajectories in behavior cloning produces a trajectory that collides directly with the obstacle. (b). Two obstacles (red and blue) partition the free space into three feasible forward directions. The tangent lines \(\mathcal{H}_1, \mathcal{H}_2\) and the observation-window edges (dashed) give three candidate modes \(m_1,m_2,m_3\).}}
\label{fig:fig_1}
\vspace{-4mm}
\end{figure}

Learning navigation policies from offline demonstrations needs neither the online interaction nor the reward design of reinforcement learning (RL)~\cite{xie2023drl, peng2025reinforcement}, and it lets a policy that sees only a local occupancy map imitate an expert planner that had access to the full map. Its central difficulty is multimodality, because a single observation often admits several valid actions, such as passing an obstacle on the left or on the right. Behavior cloning (BC)~\cite{pomerleau1988alvinn, bojarski2016end} regresses the demonstrated actions, and offline RL methods such as Decision Transformer (DT)~\cite{chen2021decision, zhang2025uncertainty} condition this regression on the desired return. Both predict one action per observation, so the regression optimum is the mean of the valid actions~\cite{shafiullah2022behavior}, and the mean of a left and a right detour points into the obstacle (\figref{fig:scenarios}).

Multimodal policy classes avoid this averaging by modeling the distribution of actions given the observation, so that a sampled action follows one alternative instead of falling between them. Behavior Transformers (BeT)~\cite{shafiullah2022behavior, lee2024behavior} learn a discrete vocabulary of actions from the demonstrations, by $k$-means or vector quantization, and predict a distribution over it together with a continuous offset. Diffusion Policy (DP)~\cite{chi2025diffusion} and its navigation variants~\cite{sridhar2024nomad} generate actions by iterative denoising, and implicit BC~\cite{florence2022implicit} selects them by minimizing a learned energy. In all of these, the alternatives are learned from the statistics of the demonstrations. The model must infer from data coverage how many alternatives a scene admits and where they lie, although in navigation both change with every obstacle arrangement. The current actions do not always reveal them either, because routes that differ in navigation demonstrations can have overlapping current actions. Moreover, a sampled action is not tied to a route around an obstacle, so successive predictions can switch routes.

In navigation, however, the alternatives have an identifiable geometric cause. Nearby obstacles split the free space into a few distinct passages (\figref{fig:modeling}), a local analogue of the homotopy classes that topological planners search over given a map~\cite{bhattacharya2012topological}. These passages can be computed from a single observation, without data or search. Computing them does not solve navigation: geometry does not indicate which passage leads to the goal beyond the observed region, it recomputes the passages at every step with no memory of the previous choice, and it gives each passage only a coarse direction rather than an executable command. The demonstrations contain what geometry lacks, because the planner that produced them saw the whole map and followed one route from start to goal.

% \begin{figure}
% \centering
% % \vspace{2mm}
%     \includegraphics[trim={0.5cm 0cm 0.5cm 0cm},clip,width=\columnwidth]{Figures/figure1.pdf}
% \caption{Two feasible avoidance trajectories pass the obstacle on opposite sides. Averaging both in behavior cloning produces a trajectory that collides directly with the obstacle.}
% \label{fig:scenarios}
% \vspace{-3mm}
% \end{figure}

\begin{figure*}
\centering
\vspace{2mm}
    \includegraphics[trim={0cm 4cm 0.7cm 4cm},clip,width=1\textwidth]{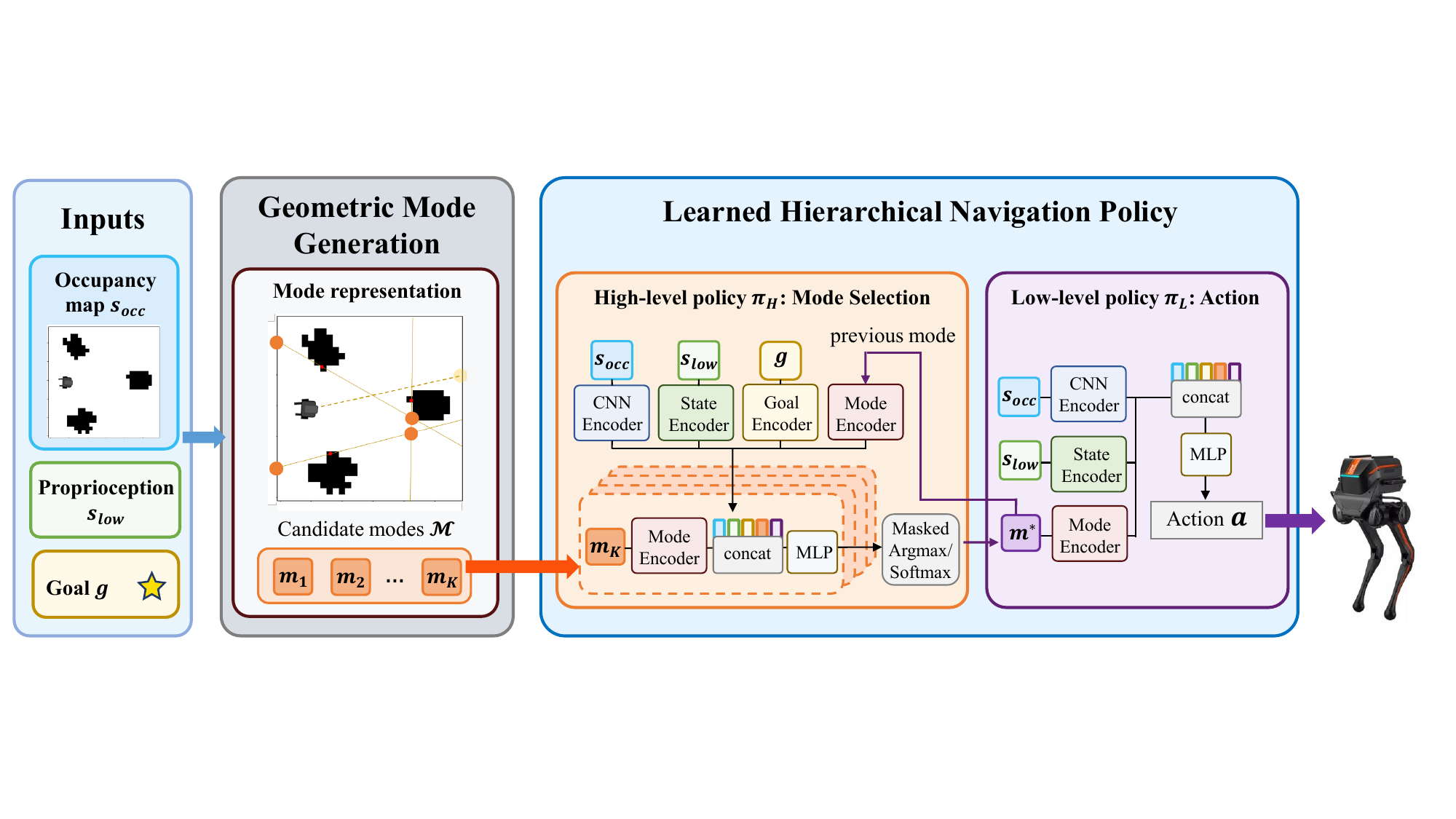}
\caption{\small{Overview of the proposed framework. The observation $s_t$ is shown as the occupancy map $s_{occ}$ and the proprioception $s_{low}$. The geometric module computes the candidate set $\mathcal{M}_t$ from $s_{occ}$ and the goal $g_t$, and is not learned. Both policies are trained from demonstrations: the high-level policy $\pi_H$ selects a mode, conditioned on $s_t$, $g_t$, and the previous mode $m_{t-1}$, and the low-level policy $\pi_L$ turns the selected mode $\hat m_t$ into a continuous action.}}
\label{fig:pipline_lite}
\vspace{-4mm}
\end{figure*}

We therefore propose a hierarchical policy in which geometry supplies the alternatives and learning selects and executes them (\figref{fig:pipline_lite}). Choosing a mode before acting has precedent: conditional imitation learning takes the mode from an external command~\cite{codevilla2018end}, and latent-plan methods learn it as a latent variable~\cite{lynch2020learning}. In our policy the modes are computed from geometry, and only the choice among them and their execution are learned. Our contributions are:
\begin{itemize}
    \item A geometric module with no learned parameters that intersects tangent lines at the closest points of nearby obstacles, with one another and with the observation-window boundary, to enumerate a variable-size set of candidate modes from a single observation.
    \item A hierarchical policy trained on demonstrations without mode annotations. A high-level policy selects a mode, conditioned on the goal and the previously selected mode; a low-level policy converts it into a continuous action and is also trained toward the mode's direction.
    \item An evaluation that separates what geometry and learning each contribute. Against four offline learning baselines, success rises from at most 43\% and 20\% in seen and unseen environments to 96.5\% and 95.2\%. A heuristic that uses the same modes without learning reaches only 79.2\% and 66.7\%, so the modes help but are not enough, and learning closes the gap with fewer mode switches and abrupt action changes.
    \item Hardware experiments in which the simulation-trained policy runs on a LimX Tron 1 bipedal robot without retraining and without tuning for each environment. The robot computes the modes from the occupancy map that it builds onboard, and navigates indoor environments with different complexities.
\end{itemize}

The rest of the paper is organized as follows. Sec.~\ref{sec:geometry} defines the problem and constructs the candidate modes from local geometry, without learning. Sec.~\ref{sec:policy} presents the hierarchical policy and how it is trained from demonstrations. Sec.~\ref{sec:exp} reports simulation and hardware experiments, and Sec.~\ref{sec:conclusion} concludes the paper.

\section{Geometric Representation of Navigation Alternatives}
\label{sec:geometry}
% Claim of this section: local geometry enumerates the alternatives. No learning
% and no dataset appear here -- everything is computable at inference time from a
% single observation. Anything that needs demonstrations or trained weights
% belongs in Sec. III (sections/3_policy.tex).

We consider goal-conditioned navigation among static obstacles, where at step $t$ the robot observes $s_t$, consisting of a local occupancy map and low-dimensional proprioception, together with a goal $g_t$ expressed in the robot frame. It outputs an action $a_t=[d_t,\theta_t]$, where $d_t$ is the desired forward displacement and $\theta_t$ the heading direction relative to the robot. A planner based on model predictive control (MPC)~\cite{peng2025real} converts this target into a velocity command, faster when $d_t$ is large and slower when it is small, and a learned gait controller tracks the command. The policy is learned from an offline dataset $\mathcal{D}=\{(s_i,g_i,a_i)\}_{i=1}^{N}$ of expert demonstrations, without interacting with the environment. This section computes a set of candidate modes $\mathcal{M}_t$ from the current observation $s_t$ and goal $g_t$ alone, without using $\mathcal{D}$. Our hierarchical policy, described in Sec.~\ref{sec:policy}, learns from $\mathcal{D}$ which mode to follow and how.

\subsection{Candidate Modes}
\label{subsec:modes}
We find the alternatives with tangent lines at the closest obstacle points, a construction commonly used in safe navigation~\cite{bohorquez2016safe, peng2025real, levy2015extended}. Let $p\in\mathbb{R}^2$ be the robot position in the robot frame. For each convex obstacle $\mathcal{O}_j$ in the observation window, let $p_j^{cl}$ be its boundary point closest to the robot. The tangent line at this point is
\begin{equation}
\label{eq:tangent}
\mathcal{H}_j
=
\left\{
x \in \mathbb{R}^2
\mid
n_j^\top (x-p_j^{cl}) = 0
\right\},
\end{equation}
where $n_j=(p-p_j^{cl})/\|p-p_j^{cl}\|_2$ is the unit vector from $p_j^{cl}$ toward the robot. The line separates the robot from the obstacle near $p_j^{cl}$, and its safe side is the half-plane that contains the robot, $\mathcal{S}_j=\{x \in \mathbb{R}^2 \mid n_j^\top (x-p_j^{cl}) \ge 0\}$.

The local observation window is a 3\,m $\times$ 3\,m square centered on the robot. We chose this size by hand, so that the window covers the area around the robot in which the perception is reliable. We compute the intersection points of each pair of lines, using all tangent lines and the four window edges. We keep an intersection point only if it lies inside the window, is not a corner of the window, and lies on the safe side $\mathcal{S}_j$ of every tangent line. Each kept intersection point defines a candidate mode $m_i$, or mode for short (\figref{fig:modeling} shows an example with three modes). We also add a goal mode when the goal direction is not blocked. The candidate set $\mathcal{M}_t$ contains all candidate modes at step $t$, and their number $K_t=|\mathcal{M}_t|$ changes with the obstacle layout.

\begin{figure}
\centering
    \includegraphics[trim={0cm 0cm 0cm 0.5cm},clip,width=1\columnwidth]{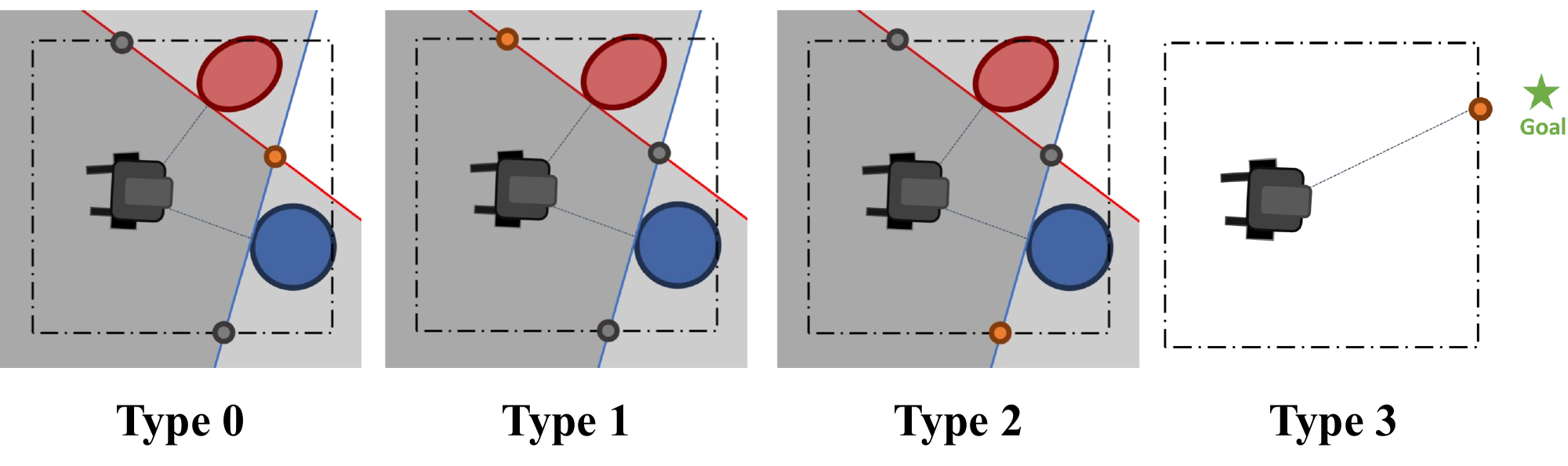}
\caption{\small{Examples of the four mode types, highlighted by orange circles. Type~0 lies between two obstacles. Types~1 and~2 pass one obstacle on the left and on the right. Type~3 points toward the goal.}}
\label{fig:mode_types}
\vspace{-4mm}
\end{figure}

Modes differ in how they are generated, and we distinguish four types (\figref{fig:mode_types}). A Type-0 mode is the intersection point of the tangent lines of two obstacles; it represents the passage between them. Type-1 and Type-2 modes are intersection points of one tangent line with a window edge; they represent passing that obstacle on the left (Type~1) or on the right (Type~2). The Type-3 mode is the goal itself if the goal lies inside the window; otherwise, it is the point where the goal direction meets the window edge. It is added only when the goal direction is not blocked, that is, when this point lies on the safe side $\mathcal{S}_j$ of every tangent line. The robot also lies on the safe side of every line, so it can then move straight to the point without leaving the safe space.
For some obstacle layouts, a Type-0 intersection point lies outside the passage between its two obstacles. We therefore add a binary passage indicator: 1 if the point lies inside the passage, and 0 otherwise. The point is inside the passage if, seen from the robot, one obstacle lies on its left and the other on its right. In other words, going counterclockwise around the robot, we meet one obstacle, then the point, then the other obstacle. For the other types, the indicator is always 1.

% Fig. 4 (Figures/stable_type.pdf) removed: it illustrated a claim that a Type-0
% mode moves into the passage over time, which no experiment supports
% (revision plan, Sec. 5). Restore from git if needed.

\subsection{Mode Encoding and Reference Action}
\label{subsec:encoding}
The point of a mode alone does not say which obstacles form the passage. The policies also see the occupancy map, but they would have to learn this link from data. We therefore give the link directly, and describe each mode $m_i$ by a vector of 22 numbers,
\begin{equation}
\label{eq:encoding}
m_i
=
\left[
m_i^{\mathrm{pt}},\,
m_i^{\mathrm{obs}},\,
m_i^{\mathrm{goal}}
\right] \in \mathbb{R}^{22}.
\end{equation}
Because all mode vectors have the same length, the candidate set can be read as a list of vectors.

The point features \(m_i^{\mathrm{pt}}=[x_i,y_i,l_i,\sin\phi_i,\cos\phi_i]\) give the point \(q_i=(x_i,y_i)\) in the robot frame, its distance \(l_i\) from the robot, and its direction \(\phi_i\). The obstacle features \(m_i^{\mathrm{obs}}\) (15 numbers) contain the mode type; for the adjacent obstacle on the left and on the right, the position, distance, and direction of its closest point, \([p_x,p_y,l,\sin\phi,\cos\phi]\); two masks that mark whether each of these obstacles exists; the passage indicator; and an initialization flag. The flag marks whether the previous-mode input is valid, because at the first step no previous mode exists and the previous-mode vector is set to zeros. 
% \placeholder{Inconsistent: the previous draft said this flag marks modes built from a window edge. Which is correct? If it marks the previous mode, what is its value for the current candidate modes (always 1)? Also, the reason given in the new draft, that an all-zero vector may also be a valid mode, does not hold: $\sin\phi_i$ and $\cos\phi_i$ cannot both be 0.} 
The two goal features $m_i^{\mathrm{goal}}$ are the distance and direction of the goal in the robot frame, and they are the same for every mode.

Each mode \(m_i\) also gives a reference action \(a_i^{\mathrm{mode}}\), which has the same form as \(a_t\) and moves the robot toward the point \(q_i\). Its heading is the direction \(\phi_i\) of the point, and its displacement is the point distance \(l_i\), limited to at most 1.5\,m. Like the modes, it is computed from geometry alone.

\section{Hierarchical Policy Learning}
\label{sec:policy}
% Claim of this section: learning decides which alternative, and how to execute it.
% Anything that requires the demonstrations or the learned parameters lives here
% (including K_max, which is chosen from dataset statistics).

The candidate set lists the alternatives at each step, but it does not say which one to take. A simple rule takes the mode whose point is closest to the goal. This rule has three gaps: it cannot tell which passage leads to the goal beyond the observation window, it has no memory of its previous choice, and it only points the robot toward a mode point, so its actions can change abruptly. The demonstrations can fill these gaps, because the planner that produced them saw the whole map and followed one route to the goal. We therefore learn a hierarchical policy from the demonstrations (\figref{fig:pipline_lite}): a high-level policy selects a mode, and a low-level policy turns the selected mode into an action.

\subsection{Demonstrations and Mode Labels}
\label{subsec:target}
\begin{figure}
\centering
\includegraphics[trim={0.2cm 1.8cm 1cm 2cm},clip,width=\columnwidth]{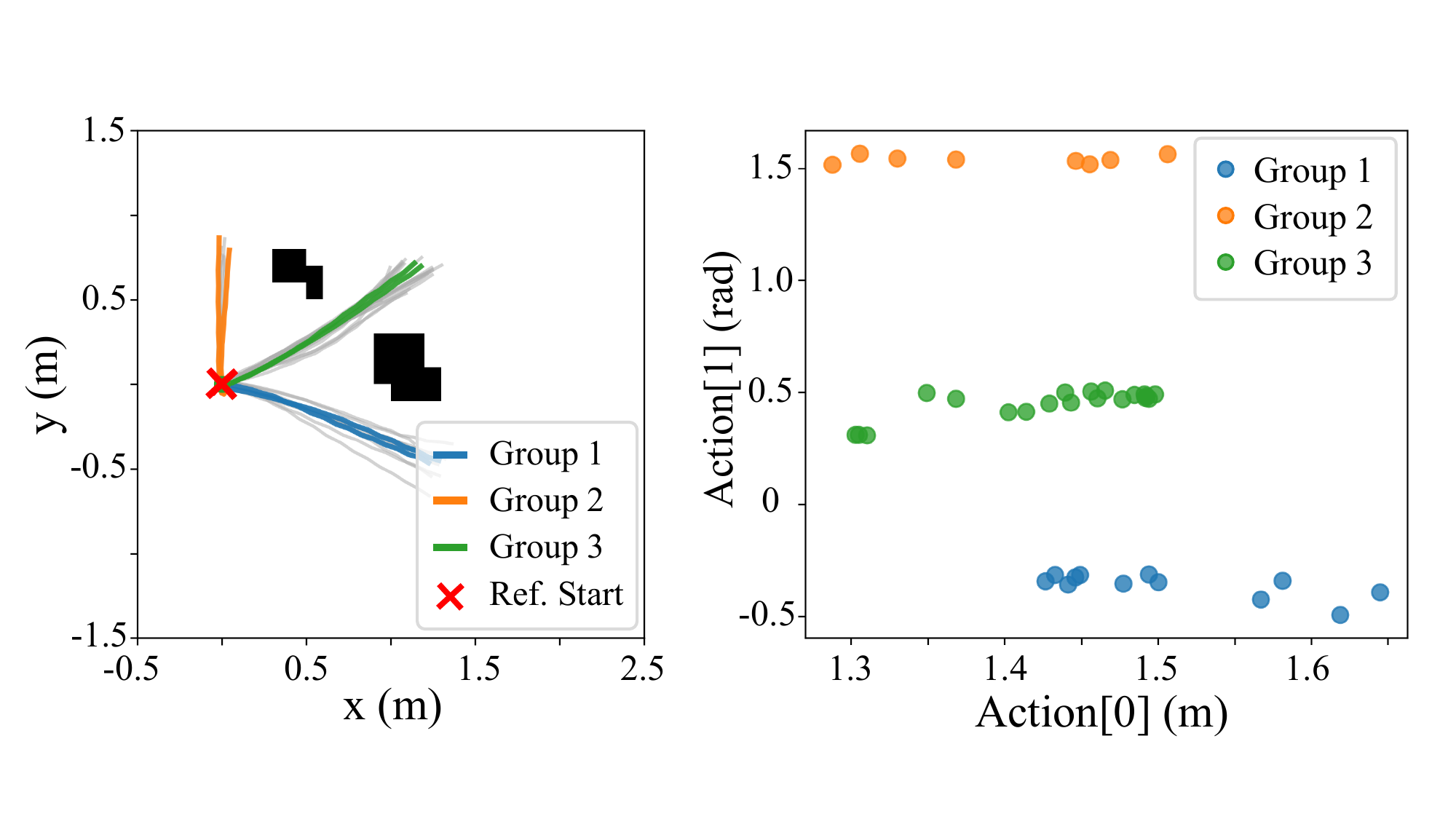}
\caption{\small{An early branching case. Demonstrations that start from similar states follow distinct future paths (left), and their current actions form separate groups (right). For display, the paths show the next 20 steps of each demonstration, and colors mark the groups that our mode labeling method finds over these 20 steps.}}
\label{fig:data_initial}
\vspace{-4mm}
\end{figure}

\begin{figure}
\centering
\includegraphics[trim={0.2cm 1.8cm 0.9cm 2cm},clip,width=\columnwidth]{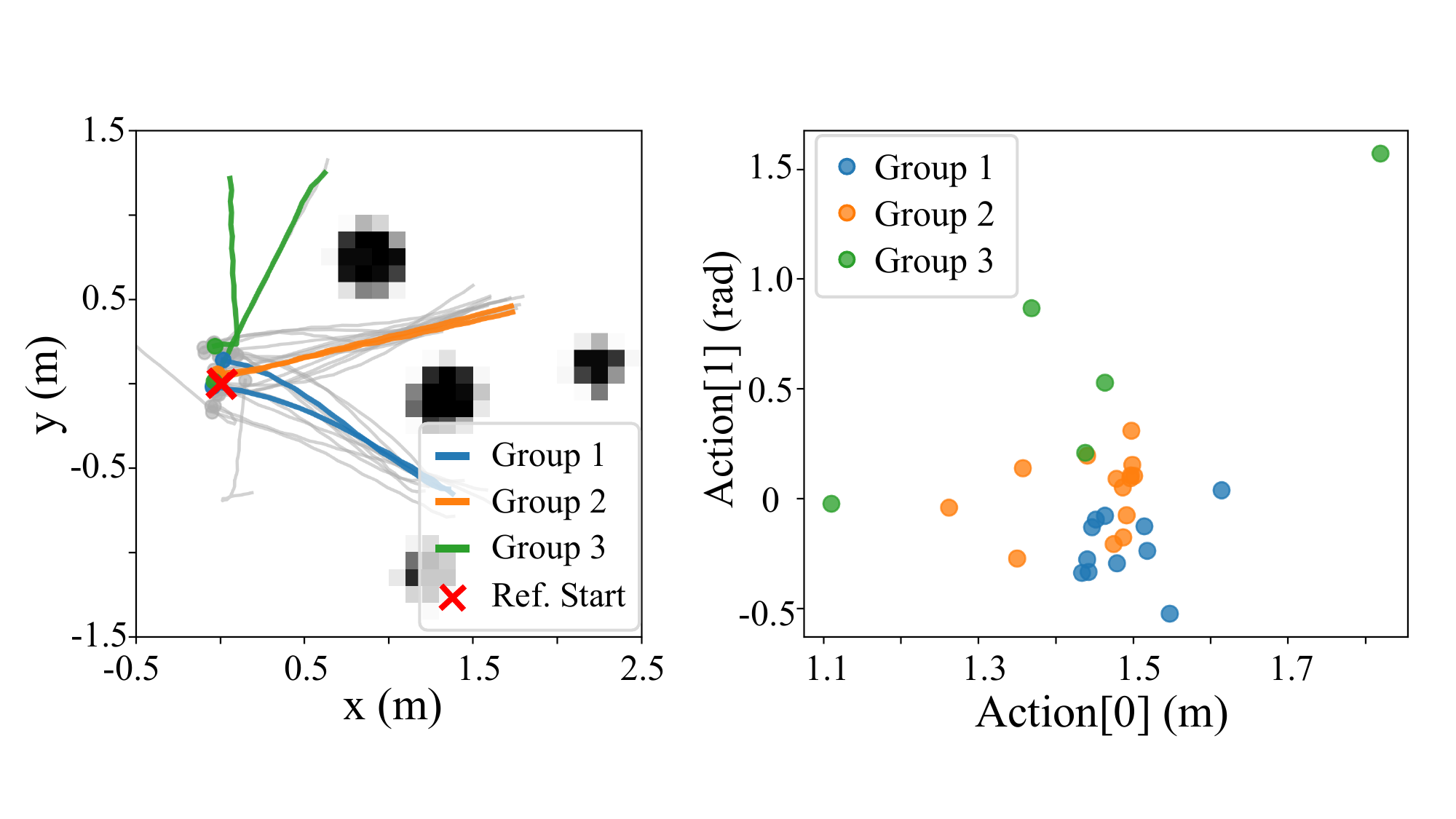}
\caption{\small{An intermediate case. Demonstrations from similar states follow distinct future paths (left), but their current actions overlap and vary within each group (right). Colors as in \figref{fig:data_initial}.}}
\label{fig:data_intermediate}
\vspace{-4mm}
\end{figure}

Demonstrations are collected in a \(5\,\mathrm{m}\times5\,\mathrm{m}\) workspace with randomly generated circular, elliptical, and polygonal obstacles. An RRT-based planner~\cite{lavalle2001rapidly} that sees the whole map generates about 10,000 trajectories of approximately 130 walking steps each, where one walking step lasts \(0.25\,\mathrm{s}\) and gives one state-action pair. Figs.~\ref{fig:data_initial} and~\ref{fig:data_intermediate} show two examples in which similar states lead to different routes. Near the start, the routes can have distinct current actions (\figref{fig:data_initial}). Later in a trajectory, the routes can differ even when their current actions overlap and vary widely (\figref{fig:data_intermediate}). Methods that learn the alternatives from the current actions may not separate such routes, whether they cluster the actions, as BeT does, or model their distribution, as DP does (Sec.~\ref{sec:intro}). We therefore define mode labels from future motion and obstacle geometry.

The demonstrations do not say which mode the planner followed, so we give each demonstration state a mode label by matching the modes to the robot's motion over the next $h=8$ steps. Let $\Delta p_t = p_{t+h} - p_t$ be the demonstrated displacement and $\Delta \psi_t = \psi_{t+h} - \psi_t$ the change of the robot heading $\psi$. The matching cost of mode $m_i$ is
\begin{equation}
\label{eq:matchcost}
C(m_i,\Delta p_t,\Delta \psi_t) = w_1 C_{\mathrm{disp}}
+
w_2 C_{\mathrm{head}}
+
w_3 C_{\mathrm{region}},
\end{equation}
where $C_{\mathrm{disp}}$ is small when the demonstrated displacement $\Delta p_t$ brings the robot closer to the point $q_i$ of the mode, and $C_{\mathrm{head}}$ is the normalized difference between the heading change $\Delta\psi_t$ and the direction $\phi_i$ of the point. $C_{\mathrm{region}}$ checks whether the future motion stays in the region of the mode, an angular sector around the robot. For Types 0--2, the sector is bounded by the rays from the robot toward the closest points of the adjacent obstacles; for Type~3, it is a cone of $\pm 30^\circ$ around the direction $\phi_i$. $C_{\mathrm{region}}$ is zero when the future motion lies inside the sector and grows with its angle to the nearest boundary of the sector. We set $(w_1,w_2,w_3)=(0.3,0.3,0.4)$ by hand, with slightly more weight on the region cost, and use the same weights for all demonstrations. The mode label is the mode with the lowest cost:
\begin{equation}
\label{eq:modelabel}
m_t^* =\arg\min_{m_i\in\mathcal{M}_t}
C(m_i,\Delta p_t,\Delta \psi_t).
\end{equation}
The labels need no manual annotation, and they are used only in training.

\subsection{Policy Architecture and Losses}
\label{subsec:architecture}
% \newsec{Policy decomposition.}
A policy $\pi(a \mid s, g)$ trained by BC minimizes
\[
\mathbb{E}_{(s,g,a)\sim\mathcal{D}}\big[\|\pi(s,g)-a\|^{2}\big],
\]
whose minimizer is the conditional mean of the demonstrated actions, which averages across modes (\figref{fig:scenarios}). We instead split each decision into two steps~\cite{le2018hierarchical}: first select a mode $m\in\mathcal{M}_t$, then choose an action for it (\figref{fig:pipline_lite}). The resulting policy is
\begin{equation}
\label{eq:policy}
\begin{aligned}
    &\pi(a \mid s_t, g_t, m_{t-1})
=\\
    &\sum_{m \in \mathcal{M}_t}
\pi_L(a \mid s_t, m)\,
\pi_H(m \mid s_t, \mathcal{M}_t, g_t, m_{t-1}),
\end{aligned}
\end{equation}
where the high-level policy $\pi_H$ gives a probability for each mode, the low-level policy $\pi_L$ gives the action for one mode, and $m_{t-1}$ is the mode selected at the previous step. Eq.~\eqref{eq:policy} sums over all modes, but at run time we use only the most likely one, the selected mode $\hat m_t=\arg\max_{m \in \mathcal{M}_t} \pi_H(m \mid s_t, \mathcal{M}_t, g_t, \hat m_{t-1})$, and the robot executes the action $\pi_L(s_t,\hat m_t)$.

\newsec{Fixed-size mode list.}
The number of modes $K_t$ changes from step to step, but batched training needs inputs of a fixed size. We therefore pad the mode list to $K_{\max}=12$ vectors with zeros, and a mask marks the padded places as invalid. $K_t\le K_{\max}$ holds for every state in the dataset and every step in the tests.

\newsec{High-level policy.}
$\pi_H$ scores each valid mode using the occupancy map, the robot state, the goal $g_t$, and the previous mode $m_{t-1}$ (\figref{fig:pipline_lite}). A CNN encodes the occupancy map, and separate encoders embed the robot state, the goal, the previous mode, and the mode vectors. For each mode, these features are concatenated and passed through a scoring network that is shared by all modes. A softmax over the valid modes turns the scores into the probabilities $\pi_H(m \mid s_t, \mathcal{M}_t, g_t, m_{t-1})$. During training, $m_{t-1}$ is the previous mode label $m_{t-1}^*$; at the first step, where no previous mode exists, it is an all-zero vector. $\pi_H$ is trained with the cross-entropy loss
\begin{equation}
\label{eq:highloss}
\mathcal{L}_{\mathrm{high}} = - \log \pi_H(m_t^* \mid s_t, \mathcal{M}_t, g_t, m_{t-1}^*).
\end{equation}
The previous mode gives $\pi_H$ a memory of its last choice, to help it keep the same choice over consecutive steps.

\newsec{Low-level policy.}
$\pi_L$ receives the current observation and one mode, and predicts a continuous action. It does not receive the goal $g_t$ directly; the goal features of the mode (Sec.~\ref{subsec:encoding}) give it the goal. During training, it receives the mode label $m_t^*$; at run time, it receives the selected mode $\hat m_t$. Training with the label gives $\pi_L$ the mode that matches the demonstrated motion, but $\pi_L$ then does not see the selection errors of $\pi_H$ during training. The action loss is
\begin{equation}
\label{eq:actionloss}
\mathcal{L}_{\mathrm{action}}
=
\left\|
\pi_L(s_t,m_t^*)-a_t
\right\|^2.
\end{equation}

\newsec{Auxiliary loss.}
We also use the reference action $a_t^{\mathrm{mode}}$ of the mode label (Sec.~\ref{subsec:encoding}) as a second training target, with the auxiliary weight $\lambda_{\mathrm{aux}}$. This weight states how reliable $a_t^{\mathrm{mode}}$ is as a target, and its rule depends on the mode type.

For Types 0--2, $a_t^{\mathrm{mode}}$ points at the intersection point of two lines. Near obstacles, few directions are safe, so a safe demonstrated action must point roughly toward the passage that the mode represents. Far from obstacles, many directions are safe, and the intersection point depends on where the lines happen to meet. The reference action is therefore a reliable target near obstacles and a weak one far from them. Let $d_{\mathrm{obs}}$ be the smallest distance from the robot to the adjacent obstacles of the mode label. We define
\begin{equation}
\label{eq:auxweight}
\lambda_{\mathrm{aux}}
=
\begin{cases}
\operatorname{clip}
\left(
\dfrac{d_{\mathrm{far}}-d_{\mathrm{obs}}}
     {d_{\mathrm{far}}-d_{\mathrm{near}}},
0,1
\right), & \text{Types 0--2},\\[2mm]
1, & \text{Type 3},
\end{cases}
\end{equation}
where $d_{\mathrm{near}}=0.3\,\mathrm{m}$ and $d_{\mathrm{far}}=0.8\,\mathrm{m}$ are chosen by hand.
For Types 0--2, the weight grows from 0 at $d_{\mathrm{far}}$ to 1 at $d_{\mathrm{near}}$.
A Type-3 mode has no adjacent obstacle, so the distance rule would give $\lambda_{\mathrm{aux}}=0$; we use a separate rule for it. Its reference action points straight at the goal, and the mode exists only when the goal direction is not blocked (Sec.~\ref{subsec:modes}). In this case we expect the demonstrated motion to head toward the goal, so the reference action is again a reliable target, and we set $\lambda_{\mathrm{aux}}=1$.
% \placeholder{Restored the earlier text. The replaced sentence said that $\lambda_{\mathrm{aux}}$ increases ``as the selected mode approaches nearby obstacles''. That is not correct: $\lambda_{\mathrm{aux}}$ depends on the distance from the robot to the adjacent obstacles of the mode label, and it is used only in training, where there is no selected mode.}

The low-level loss adds the auxiliary loss to the action loss:
\begin{equation}
\label{eq:lowloss}
\mathcal{L}_{\mathrm{low}}
=
\mathcal{L}_{\mathrm{action}}
+
\lambda_{\mathrm{aux}}
\left\|
\pi_L(s_t,m_t^*)-a_t^{\mathrm{mode}}
\right\|^2.
\end{equation}
For one training sample, this loss is smallest when $\pi_L$ outputs the weighted average $(a_t+\lambda_{\mathrm{aux}}a_t^{\mathrm{mode}})/(1+\lambda_{\mathrm{aux}})$. Since $\lambda_{\mathrm{aux}}\le 1$, the reference action receives at most half of the weight, and the demonstrated action always keeps at least half. Even for Type 3, the target is the midpoint of the demonstrated action and the reference action, not the reference action alone.

\section{Simulation and Hardware Experiments}
\label{sec:exp}
We test the policy in simulation and on a bipedal robot, and the simulation experiments ask two questions. First, how does the policy compare with offline learning methods that do not use the modes (Sec.~\ref{subsec:baselines})? Second, how does it compare with model variants, including the Heuristic, which uses the same modes without learning (Sec.~\ref{subsec:ablation})? The hardware experiments test whether the policy, trained only in simulation, runs on a real robot with onboard perception (Sec.~\ref{subsec:hardware}).

\subsection{Experimental Setup}
\label{subsec:setup}
\newsec{Training.}
$\pi_H$ and $\pi_L$ are trained separately on the demonstrations of Sec.~\ref{subsec:target} by minimizing $\mathcal{L}_{\mathrm{high}}$ and $\mathcal{L}_{\mathrm{low}}$, respectively. We split the dataset at the trajectory level and hold out 15\% of the trajectories for validation. Both policies are optimized using AdamW with a batch size of 256. For $\pi_H$, we use a learning rate of $10^{-3}$, a weight decay of $10^{-4}$, and train for 10 epochs; for $\pi_L$, we use a learning rate of $10^{-5}$, a weight decay of $10^{-5}$, and train for 20 epochs. We repeat training with five random seeds, and report the mean and standard deviation across seeds in Tables~\ref{tab:main} and~\ref{tab:ablation}. Training one complete hierarchical policy takes approximately 1.2\,h on a single NVIDIA T1200 Laptop GPU. We monitor the training and validation losses of both policies after every epoch.

\newsec{Test cases.}
Test maps use the same workspace and obstacle shapes as the demonstrations (Sec.~\ref{subsec:target}). The seen setting has 250 test cases with obstacle counts that also appear in the demonstrations: 50 cases without obstacles, which differ only in their start and goal positions, and 100 cases each with four and eight obstacles, some with narrow gaps between them. The unseen setting has 30 new maps, 10 each with 6, 10, and 14 obstacles. The count 6 lies between the seen counts of 4 and 8, and 10 and 14 are larger than any seen count, so these maps test new layouts and more obstacles. Their gaps between obstacles, however, are wider than the narrowest gaps in the seen maps.

\newsec{Trials.}
All methods are tested on the same cases, with the same start and goal positions. Each method runs in closed loop, receiving at every step the observation that results from its own previous actions and outputting the next action, so its errors add up over the trial. A step is one action of the policy, which lasts one walking step of $0.25\,$s (Sec.~\ref{subsec:target}). The simulator executes each action with the MPC planner and the learned gait controller of Sec.~\ref{sec:geometry}. A trial ends when the robot reaches within $0.15\,$m of the goal, collides with an obstacle, meaning that its center comes within $0.2\,$m of one, or reaches the step limit of 200 steps. We report the success rate (SR), collision rate (CR), and timeout rate (TR), with $\mathrm{SR}+\mathrm{CR}+\mathrm{TR}=100\%$.

\newsec{Efficiency and smoothness.}
The path ratio (Path R.) is the total path length of a method divided by that of Ours. The step ratio (Step R.) is the same ratio for the number of steps. Both sums use only the cases that the method and Ours both complete, so the two are compared on the same cases. Values below one mean shorter paths or fewer steps than Ours. Since every step has the same duration, fewer steps means that the robot reaches the goal sooner. An action jump is a step at which the heading $\theta_t$ changes by more than $\pi/6$ from the previous step, and a mode switch is a step at which the selected mode differs from the previous one. A large turn is a step at which the commanded direction differs from the current heading of the robot by more than $45^\circ$; it measures how far the robot has to turn, not how much the command changes. Action jumps and mode switches are counted over all trials, including collisions and timeouts, and divided by the total number of steps, which gives counts per 100 steps. We also report the mean computation time per step, measured on the laptop used for training.

\subsection{Comparison with Offline Learning Baselines}
\label{subsec:baselines}

\begin{table*}[t]
\centering
\small
\setlength{\tabcolsep}{6pt}
\begin{tabular}{l|llllllll}
\toprule
\textbf{ } & \textbf{Method} & \textbf{SR (\%)}$\uparrow$ & \textbf{CR (\%)}$\downarrow$ & \textbf{TR (\%)}$\downarrow$ & \textbf{Path R.}$\downarrow$ & \textbf{Step R.}$\downarrow$ & \textbf{\makecell[l]{Act. Jump\\(\#/100)}}$\downarrow$ & \textbf{\makecell[l]{Comput.\\ Time (ms)}}$\downarrow$ \\
\midrule

\multirow{5}{1em}{\rotatebox{90}{\textbf{Seen}}}
& MLP-BC & 42.8$\pm$5.9 & 56.0$\pm$6.0 & 1.2$\pm$2.0 & 1.04 & 1.09 & 0 & 1.2 \\
& DT     & 29.1$\pm$6.8 & 70.9$\pm$6.8 & 0 & 0.94 & 1.18 & 0.08$\pm$0.07 & 3.0 \\
& BeT    & 34.8$\pm$1.8 & 65.0$\pm$2.0 & 0.2$\pm$0.3 & 0.97 & 1.04 & 1.50$\pm$0.60 & 3.6 \\
& DP     & 42.7$\pm$3.2 & 55.2$\pm$4.2 & 2.1$\pm$1.7 & 1.12 & 1.18 & 3.38$\pm$0.20 & 98.7 \\
& \textbf{Ours}   & \textbf{96.5$\pm$1.0} & \textbf{3.2$\pm$1.0} & 0.3$\pm$0.2 & 1.00 & 1.00 & 0.25$\pm$0.02 & 2.9 \\

\midrule

\multirow{5}{1em}{\rotatebox{90}{\textbf{Unseen}}}
& MLP-BC & 20.0$\pm$7.3 & 80.0$\pm$7.3 & 0 & 0.88 & 0.91 & 0 & 1.4 \\
& DT     & 15.3$\pm$3.4 & 84.7$\pm$3.4 & 0 & 0.94 & 1.17 & 0.22$\pm$0.36 & 3.4 \\
& BeT    & 20.0$\pm$6.0 & 80.0$\pm$6.0 & 0 & 0.89 & 1.01 & 1.80$\pm$0.71 & 3.4 \\
& DP     & 18.7$\pm$6.9 & 81.3$\pm$6.9 & 0 & 1.01 & 1.05 & 3.24$\pm$0.20 & 100.5 \\
& \textbf{Ours}   & \textbf{95.2$\pm$4.5} & \textbf{2.6$\pm$2.6} & 2.2$\pm$1.4 & 1.00 & 1.00 & 0.51$\pm$0.12 & 3.4 \\

\bottomrule
\end{tabular}
\caption{Comparison with offline learning baselines. Path R. and Step R. are relative to Ours, on the cases that both methods complete. Mean $\pm$ standard deviation over five training seeds (Sec.~\ref{subsec:setup}).}
\label{tab:main}
\vspace{-0.5em}
\end{table*}

We compare Ours with the four offline learning methods of Sec.~\ref{sec:intro}: BC with a multilayer perceptron (MLP-BC), DT, BeT, and DP, none of which uses the modes. All are trained on the same demonstrations and receive the same inputs as Ours: the occupancy map, the robot state, and the goal.
% \placeholder{Implementations and network sizes.}

Table~\ref{tab:main} shows that Ours succeeds in 96.5\% of seen and 95.2\% of unseen cases, while no baseline exceeds 43\% and 20\%. Almost all baseline failures are collisions. Their timeout rates are at most 2.1\%, while on unseen maps their collision rates are 80\% or higher, compared with 2.6\% for Ours. The results are consistent with the two failure types described in Sec.~\ref{sec:intro}. MLP-BC and DT predict one action per observation. Their actions are smooth, with at most 0.22 action jumps per 100 steps, yet they collide in 56--85\% of trials, as expected from actions that average between alternatives. BeT and DP model several possible actions. They make 1.50--3.38 action jumps per 100 steps, compared with 0.25 and 0.51 for Ours, and they still collide in 55--81\% of trials, as expected from predictions that switch between alternatives. The demonstrations make these alternatives hard to learn from data, because routes that differ can have overlapping current actions (\figref{fig:data_intermediate}). Ours first selects one mode and then predicts an action for that mode only, which avoids both failure types. These explanations fit the numbers, but they are not proven for each collision. Still, the large gap in success shows that structuring the decision with the geometric modes makes navigation much more robust. With many more demonstrations, or with inputs that show more of the map, BeT and DP may separate the routes better. This is outside the scope of this paper, which compares all methods on the same demonstrations and test cases.

The path and step ratios cover only the cases that a baseline also completes, which are at most 43\% of seen and 20\% of unseen cases. On these cases, the path lengths of the baselines differ from those of Ours by at most 12\% (path ratio 0.88--1.12). Most baselines need more steps than Ours (step ratio 1.01--1.18). The exception is MLP-BC on unseen maps, which needs fewer steps (0.91) on the 20\% of cases it completes. Overall, Ours has few action jumps and competitive path lengths, needs fewer steps than most baselines, and succeeds far more often.

Ours needs 2.9\,ms per step on seen maps and 3.4\,ms on unseen maps, similar to MLP-BC, DT, and BeT (1.2--3.6\,ms). DP needs about 100\,ms, because it generates each action by iterative denoising. Ours thus runs in real time, well within the 0.25\,s of one step, so the hierarchy adds robustness without losing the real-time computation needed for closed-loop control on a robot.

\subsection{Comparison with Model Variants}
\label{subsec:ablation}

\begin{table*}[t]
\centering
\small
\setlength{\tabcolsep}{6pt}

\begin{tabular}{l|lllllllll}
\toprule
\textbf{ } & \textbf{Method} & \textbf{SR (\%)}$\uparrow$ & \textbf{CR (\%)}$\downarrow$ & \textbf{TR (\%)}$\downarrow$ & \textbf{Path R.}$\downarrow$ & \textbf{Step R.}$\downarrow$ & \textbf{\makecell[l]{Act. Jump\\(\#/100)}}$\downarrow$ & \textbf{\makecell[l]{Comput.\\ Time (ms)}}$\downarrow$ & \textbf{\makecell[l]{Mode Sw.\\ (\#/100)}}$\downarrow$\\
\midrule

\multirow{5}{1em}{\rotatebox{90}{\textbf{Seen}}}
& w/o Auxi.  & 70.4$\pm$12.4 & 25.8$\pm$11.3 & 3.8$\pm$2.2 & 0.95 & 0.94 & 0 & 2.6 & 6.05$\pm$0.41 \\
& w/o Hier.  & 25.8$\pm$15.5 & 60.3$\pm$4.7 & 13.8$\pm$15.3 & 1.27 & 1.17 & 0 & 2.4 & N/A \\
& Mode-act.-only  & 97.8$\pm$0.7 & 0.2$\pm$0.2 & 2.0$\pm$0.6 & 1.09 & 1.10 & 0.65$\pm$0.08 & 3.0 & 5.36$\pm$0.09 \\
& Heuristic  & 79.2 & 8.4 & 12.4 & 1.01 & 1.27 & 12.14 & 2.2 & 17.79 \\
% \cmidrule(lr){2-9}
& \textbf{Ours} & 96.5$\pm$1.0 & 3.2$\pm$1.0 & 0.3$\pm$0.2 & 1.00 & 1.00 & 0.25$\pm$0.02 & 2.9 & 5.53$\pm$0.05 \\

\midrule

\multirow{5}{1em}{\rotatebox{90}{\textbf{Unseen}}}
& w/o Auxi.  & 62.0$\pm$22.0 & 30.7$\pm$21.5 & 7.3$\pm$3.9 & 0.91 & 0.90 & 0 & 3.3 & 7.41$\pm$0.77 \\
& w/o Hier.  & 10.7$\pm$7.1 & 84.0$\pm$3.9 & 5.3$\pm$5.0 & 1.54 & 1.34 & 0 & 2.6 & N/A \\
& Mode-act.-only  & 98.0$\pm$1.6 & 0 & 2.0$\pm$1.6 & 1.07 & 1.08 & 0.92$\pm$0.11 & 3.8 & 7.13$\pm$0.14 \\
& Heuristic  & 66.7 & 10.0 & 23.3 & 0.98 & 1.33 & 18.42 & 3.0 & 25.36 \\
& \textbf{Ours} & 95.2$\pm$4.5 & 2.6$\pm$2.6 & 2.2$\pm$1.4 & 1.00 & 1.00 & 0.51$\pm$0.12 & 3.4 & 7.40$\pm$0.27 \\
\bottomrule
\end{tabular}
\caption{Comparison with the model variants. Metrics as in Table~\ref{tab:main}. The Heuristic is not trained, so it has no $\pm$. w/o Hier. selects no mode, so it has no mode switches (N/A).}
\label{tab:ablation}
\vspace{-1em}
\end{table*}

We compare Ours with four variants, all using the same modes and demonstrations. They differ in how the mode is selected and in what the low-level policy learns from.
% Confirmed: w/o Hier. receives the mode vectors; MLP-BC does not.
\begin{itemize}
\item \emph{w/o Hier.}: no mode selection. A single network receives the same state, goal, and candidate modes as Ours and predicts the action directly.
% \item \emph{w/o Hier.}: no mode selection. A single network predicts the action from the observation, the goal, and the mode vectors; MLP-BC (Sec.~\ref{subsec:baselines}) does not receive the mode vectors. \placeholder{Padded list of $K_{\max}$ vectors, or a summary? Which loss?}
\item \emph{w/o Auxi.}: as Ours, but with $\lambda_{\mathrm{aux}}=0$, so $\pi_L$ learns only from the demonstrated actions.
\item \emph{Mode-act.-only}: as Ours, but $\pi_L$ learns only from the reference actions, with the loss $\|\pi_L(s_t,m_t^*)-a_t^{\mathrm{mode}}\|^2$.
% Confirmed (student, 2026-09-15): Mode-act.-only uses only the reference-action loss, with weight 1 at every step, and no demonstrated actions.
\item \emph{Heuristic}: nothing is learned. It selects the mode whose point is closest to the goal in Euclidean distance and executes the reference action of that mode (Sec.~\ref{subsec:encoding}).
% \item \emph{Heuristic}: nothing is learned. It selects the mode whose point is closest to the goal and sends its reference action to the robot. Confirmed (student, 2026-09-15): Euclidean distance to the goal; $\mathcal{M}_t$ is never empty; without obstacles, the Type-3 mode points to the goal.
\end{itemize}
Table~\ref{tab:ablation} shows that both mode selection and the auxiliary loss are important for robust navigation.

\newsec{Mode selection.}
w/o Hier. succeeds in 25.8\% of seen and 10.7\% of unseen cases. This is no higher than MLP-BC (42.8\% and 20.0\%, Table~\ref{tab:main}), which does not receive the mode vectors. Giving the modes to a network as extra input is therefore not enough. Every method that selects a mode first, including the Heuristic, succeeds more often than both (62.0--98.0\%). A possible reason is that a single network that sees several modes can still average between them.

\newsec{Auxiliary loss.}
The success of w/o Auxi. drops to 70.4\% and 62.0\%, mostly because of collisions (25.8\% and 30.7\%), and it varies widely across training seeds ($\pm$12.4 and $\pm$22.0). Even with a selected mode, learning only from the demonstrated actions does not keep the robot reliably clear of obstacles. This supports using the reference action as a second target near obstacles (Sec.~\ref{subsec:architecture}).

\newsec{Demonstrated actions.}
The success of Mode-act.-only is similar to or slightly higher than that of Ours (97.8\% and 98.0\%), and it almost never collides (0.2\% and 0\%), so the reference actions are a strong target for safety. However, its paths are 9\% and 7\% longer, it needs 10\% and 8\% more steps, and it makes more action jumps (0.65 vs.\ 0.25 and 0.92 vs.\ 0.51 per 100 steps). Learning only from the reference actions is thus safe but conservative. Learning from the demonstrated actions as well gives shorter and smoother motion, at the cost of slightly more collisions (3.2\% and 2.6\%), while success stays above 95\%.

\newsec{Heuristic.}
The Heuristic succeeds in 79.2\% of seen and 66.7\% of unseen cases. This is more than every baseline in Table~\ref{tab:main}, so the modes themselves carry useful navigation structure, but less than Ours (96.5\% and 95.2\%). The Heuristic collides in 8.4\% and 10.0\% of cases, compared with 3.2\% and 2.6\% for Ours, and it times out in 12.4\% and 23.3\%, compared with 0.3\% and 2.2\%. On the cases that both complete, its paths have similar length (path ratio 1.01 and 0.98), but it needs 27\% and 33\% more steps. It switches modes about three times as often as Ours (17.79 vs.\ 5.53 and 25.36 vs.\ 7.40 per 100 steps) and makes many more action jumps (12.14 vs.\ 0.25 and 18.42 vs.\ 0.51).

\begin{figure}
\centering
    \begin{subfigure}[b]{0.95\columnwidth}
         \centering
         \includegraphics[trim={0cm 17cm 0cm 0cm},clip,width=\textwidth]{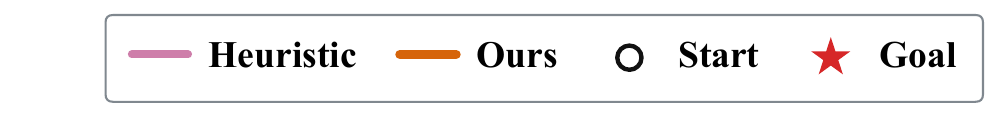}
     \end{subfigure}
     \hfill
    \begin{subfigure}[b]{0.49\columnwidth}
         \centering
         \includegraphics[trim={0.2cm 0.2cm 0.3cm 2cm},clip,width=\textwidth]{Figures/seen_035.pdf}
         \caption{}
         \label{fig:unseen_22}
     \end{subfigure}
     \hfill
     \begin{subfigure}[b]{0.49\columnwidth}
         \centering
         \includegraphics[trim={0.2cm 0.2cm 1.2cm 2cm},clip,width=\textwidth]{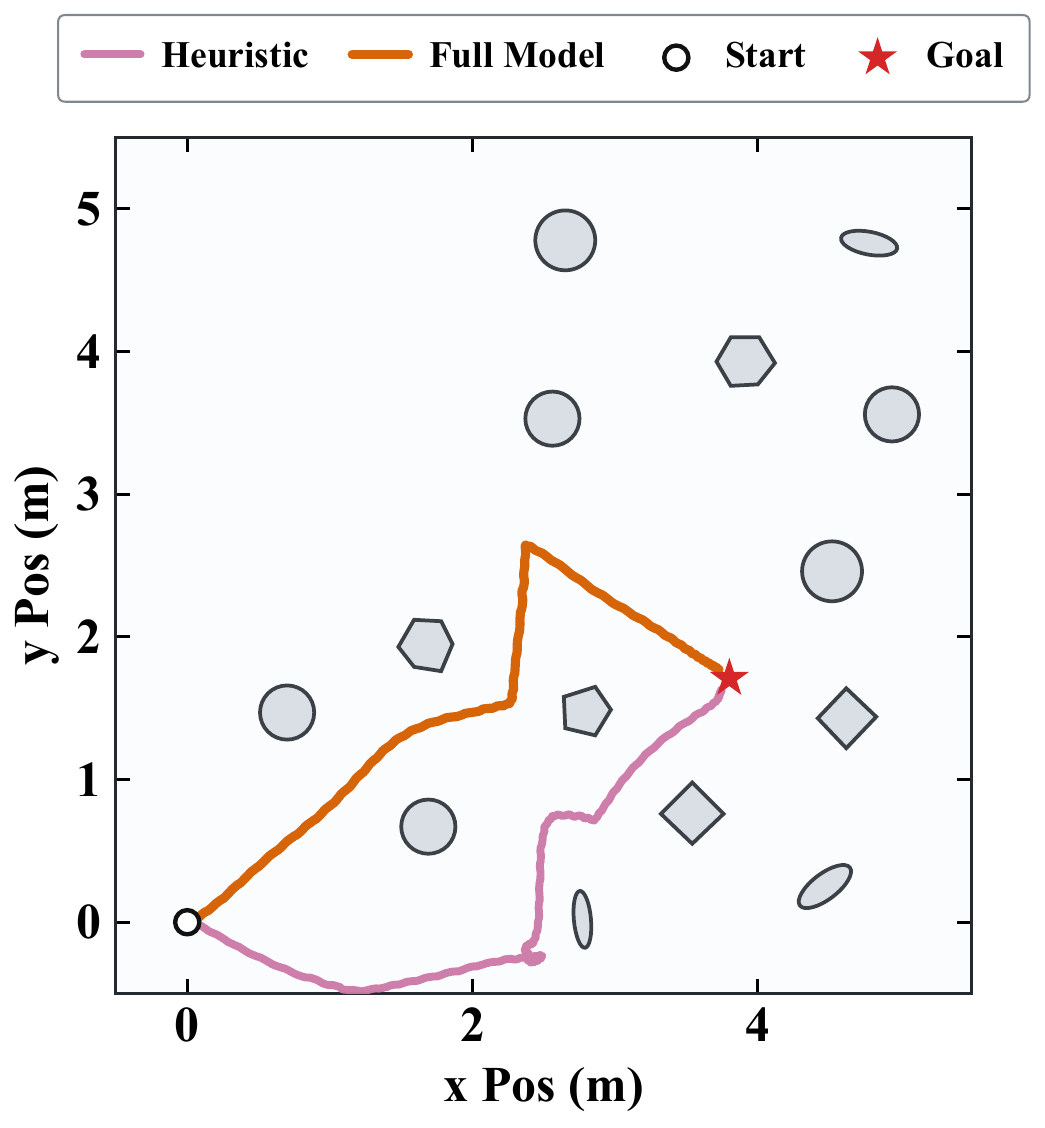}
         \caption{}
         \label{fig:unseen_24}
     \end{subfigure}
\caption{\small{Trajectories of the Heuristic and Ours in two test cases. (a) A seen case: the Heuristic collides with an obstacle ($\times$), while Ours reaches the goal. (b) An unseen case: both reach the goal by different routes.}}
\label{fig:unseen_comp}
\vspace{-4mm}
\end{figure}

\begin{table}[ht]
\vspace{2mm}
\centering
\small
\setlength{\tabcolsep}{4pt}
\renewcommand{\arraystretch}{1.05}
\begin{tabular}{lllll}
\toprule
& \multicolumn{2}{c}{Case (a)} & \multicolumn{2}{c}{Case (b)} \\
\cmidrule(lr){2-3}\cmidrule(lr){4-5}
& Heuristic & Ours & Heuristic & Ours \\
\midrule
Outcome & Collision & Success & Success & Success  \\
Steps & 108 (F) & 103 & 143 & 98 \\
Path length (m) & 5.3 (F) & 6.9 & 5.9 & 5.7 \\
Action jumps & 30 & 0 & 40 & 1 \\
Large turns & 32 & 0 & 19 & 5 \\
Mode switches & 42 & 8 & 55 & 10 \\
\bottomrule
\end{tabular}
\caption{The seen case (a) and the unseen case (b) of \figref{fig:unseen_comp}. Counts are totals over the trial. (F): failed trial.}
\label{table:heuristic_unseen_cases}
\vspace{-3mm}
\end{table}

\newsec{Two examples.}
% \figref{fig:unseen_comp} and Table~\ref{table:heuristic_unseen_cases} show two unseen cases. In case (a), both methods reach the goal by different routes with similar path lengths (5.9 and 5.7\,m). The Heuristic, however, needs 143 steps instead of 98. It also makes 40 action jumps, 19 large turns, and 55 mode switches, compared with 1, 5, and 10 for Ours. In case (b), the Heuristic switches modes 65 times, makes 49 action jumps and 51 large turns, and does not reach the goal within 200 steps. Ours reaches the goal in 115 steps. The path of the Heuristic is shorter here (5.6 vs.\ 7.4\,m) only because its trial did not finish. A likely reason for this behavior is that the Heuristic selects the closest mode again at every step. As the robot moves, the mode points move, so its choice can change and the robot turns back and forth instead of moving on. This matches two gaps named in Sec.~\ref{sec:policy}: the Heuristic has no memory of its previous choice, and it only points the robot toward a mode point. With the same modes, the learned policy succeeds more often and moves more consistently.
\figref{fig:unseen_comp} and Table~\ref{table:heuristic_unseen_cases} show one seen and one unseen case. In case (a), the Heuristic changes its direction again and again near the obstacles and collides after 108 steps. Until then, it makes 30 action jumps, 32 large turns, and 42 mode switches. Ours reaches the goal in 103 steps, with no action jump, no large turn, and 8 mode switches. In case (b), both methods reach the goal with similar path lengths (5.9 and 5.7\,m), but the Heuristic needs 143 steps instead of 98, and it makes many more action jumps (40 vs.\ 1), large turns (19 vs.\ 5), and mode switches (55 vs.\ 10). A likely reason is that the Heuristic selects the mode closest to the goal again at every step. As the robot moves, the mode points move, so its choice can change and the robot turns back and forth instead of moving on; in case (a), this brings it into an obstacle. This matches two gaps named in Sec.~\ref{sec:policy}: the Heuristic has no memory of its previous choice, and it only points the robot toward a mode point. With the same modes, the learned policy succeeds more often and moves more consistently.

\begin{figure}
\centering
\vspace{2mm}
    \includegraphics[trim={5.5cm 0.2cm 4.5cm 0cm},clip,width=1\columnwidth]{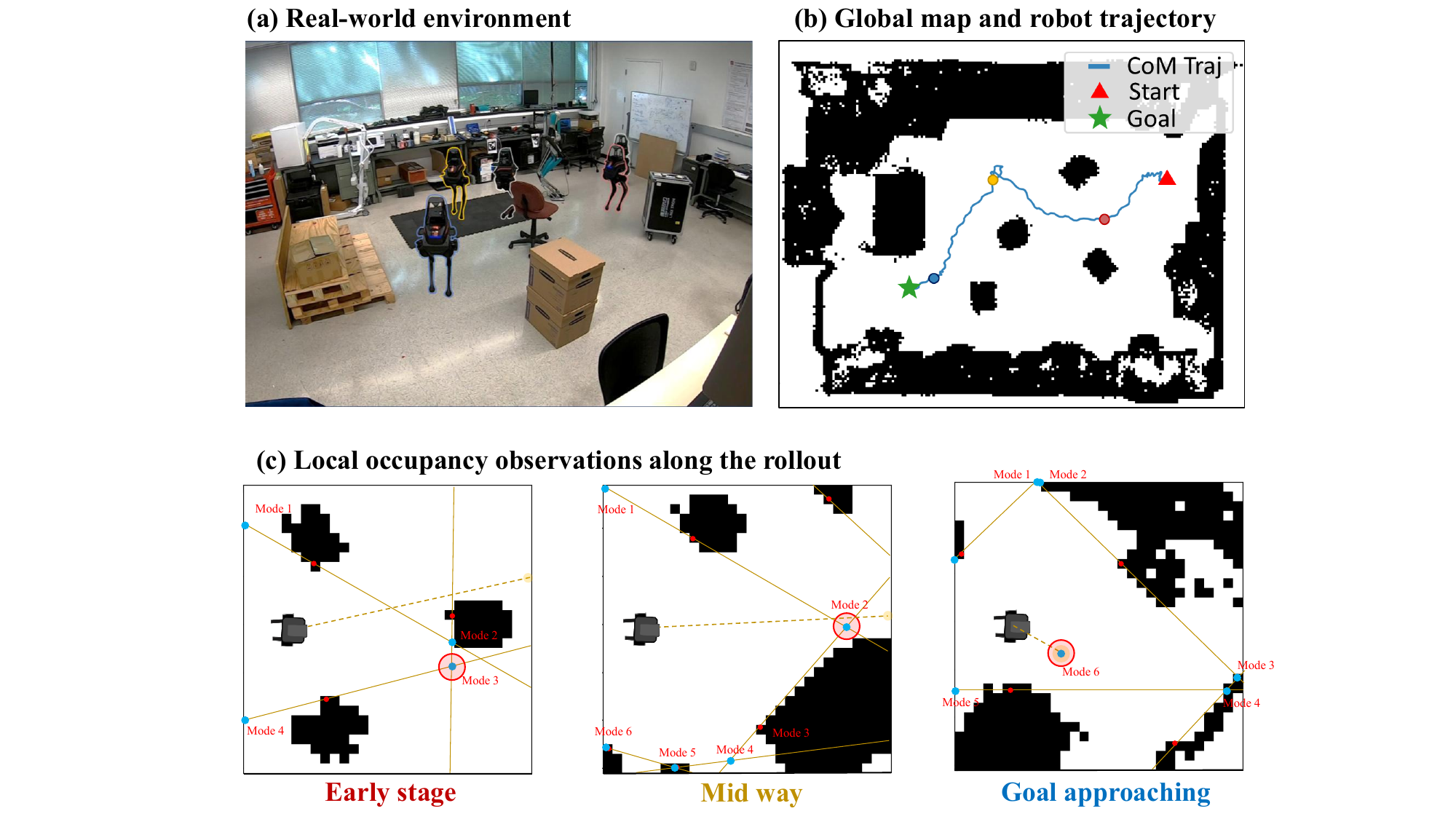}
\caption{\small{Navigation on the LimX Tron 1 robot. (a) Laboratory with static obstacles. (b) Map built by the robot's perception, with the path of the robot from start to goal. (c) Local occupancy maps at three stages of the run, with the candidate modes and the selected mode (circled); the markers in (b) show where each map was taken.}}
\label{fig:hardware}
\vspace{-4mm}
\end{figure}

\subsection{Hardware Experiments}
\label{subsec:hardware}
\begin{figure*}
\centering
\vspace{2mm}
    \includegraphics[trim={0cm 0cm 0cm 0cm},clip,width=1\textwidth]{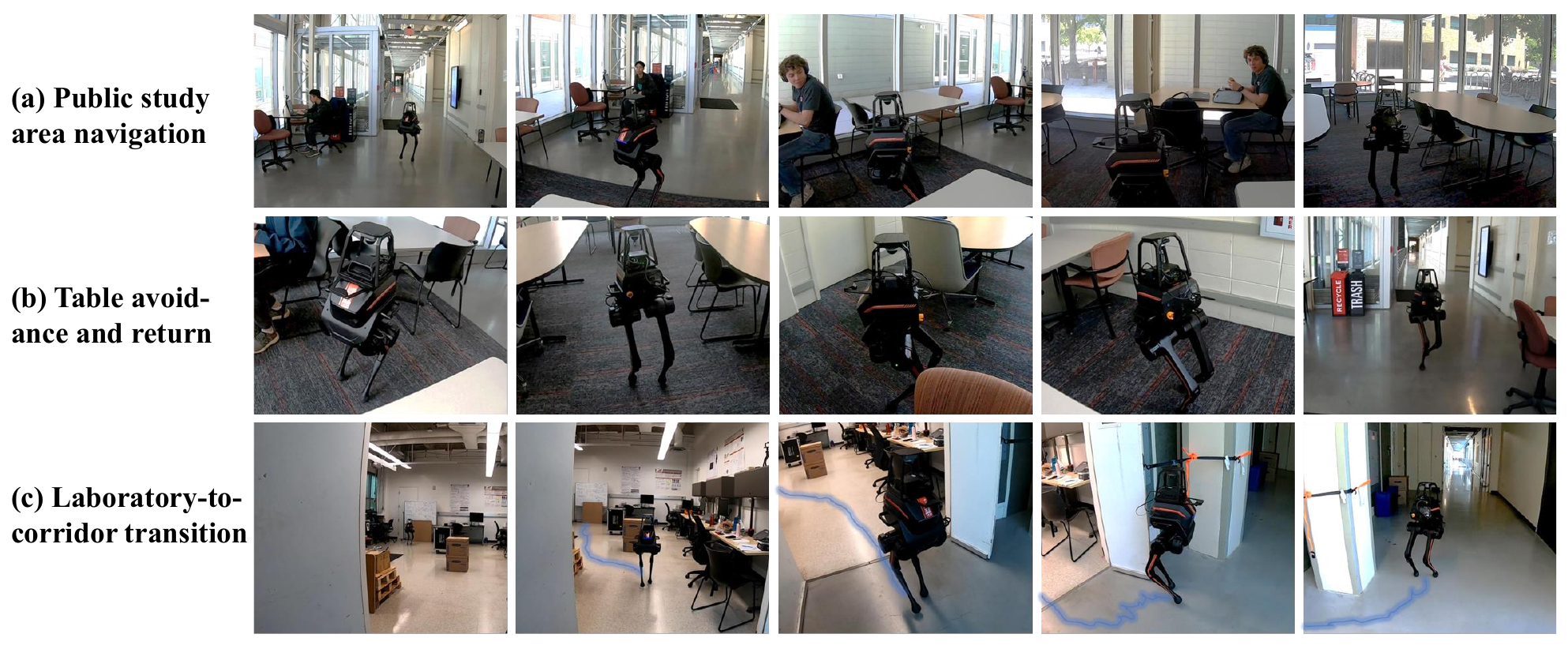}
\caption{\small{Deployments in three more indoor environments: (a) a public study area, (b) table avoidance and return, (c) a doorway from the laboratory into a corridor.}}
\label{fig:additional_nav}
\vspace{-2mm}
\end{figure*}
% We run the policy on a LimX Tron 1 bipedal robot in indoor environments with static obstacles, without retraining and without tuning for each environment. The robot builds the local occupancy map from onboard perception, and the policy runs in closed loop as in simulation. \placeholder{Sensor; how the map is built; how obstacles are extracted from it for the tangent lines; steps per second.} The real maps differ from the training data in two ways. The obstacles are furniture, boxes, and walls instead of circles, ellipses, and polygons, and the maps contain noise (\figref{fig:hardware}).
We run the policy on a LimX Tron 1 bipedal robot in indoor environments with static obstacles, without retraining and without tuning for each environment. The robot builds the local occupancy map from a Livox Mid-360 LiDAR. The modes are computed from this map, and the policy runs in closed loop at 24\,Hz. The real maps differ from the training data in two ways. The obstacles are furniture, boxes, and walls instead of circles, ellipses, and polygons, and the maps contain perception noise (\figref{fig:hardware}).

In the laboratory (\figref{fig:hardware}), the robot consistently reaches the goal, passing several obstacles and changing its direction as new obstacles enter the observation window. \figref{fig:hardware}(c) shows the candidate modes and the selected mode on three of these real local maps. \figref{fig:additional_nav} shows three more environments, each with a different difficulty: (a) a public study area with tables, chairs, and seated people; (b) a table that the robot avoids on its way out and back; and (c) a doorway from the laboratory into a corridor, which is a narrow passage followed by long walls. These deployments show that the policy, trained only in simulation, transfers to real indoor environments.

\section{Conclusion} 
\label{sec:conclusion}
We presented a hierarchical policy for offline navigation learning in which local geometry supplies the candidate modes and learning selects and executes them. In simulation, it succeeded far more often than offline learning baselines. Compared with a heuristic that uses the same modes, it succeeded more often, collided less, and switched modes less often. The policy trained in simulation also ran on a bipedal robot without retraining. Future work will extend the approach to moving obstacles.

% \section*{Acknowledgement}
% Generative AI tools were used in this work. ChatGPT (OpenAI) and Claude (Anthropic) assisted in code development and in revising the text in the paper.  All AI-assisted content was reviewed, edited, and verified by the authors.

% \newpage
\bibliographystyle{IEEETran}
\bibliography{references.bib}

\end{document}